%% file: main.tex
\documentclass[10pt,twocolumn,letterpaper]{article}

\usepackage[pagenumbers]{cvpr} % To force page numbers, e.g. for an arXiv version
\usepackage{colortbl}
\usepackage{multirow}
\usepackage{graphicx}
\usepackage{tablefootnote}
\usepackage{tabu}
\input{preamble}

\definecolor{cvprblue}{rgb}{0.21,0.49,0.74}
\usepackage[pagebackref,breaklinks,colorlinks,citecolor=cvprblue]{hyperref}

\def\paperID{8653} % *** Enter the Paper ID here
\def\confName{CVPR}
\def\confYear{2024}

\usepackage{pifont}
\newcommand{\cmark}{\ding{51}}
\newcommand{\xmark}{\ding{55}}
\newcommand{\SegNext}{\textbf{SegNext}}

\title{Rethinking Interactive Image Segmentation \\ with Low Latency, High Quality, and Diverse Prompts}

\author{
Qin Liu \hspace{3mm} Jaemin Cho \hspace{3mm} Mohit Bansal \hspace{3mm} Marc Niethammer \\
University of North Carolina at Chapel Hill \\
{\tt\small \href{https://github.com/uncbiag/SegNext}{https://github.com/uncbiag/SegNext}}
}

\begin{document}
\maketitle
\input{sec/0_abstract}    
\input{sec/1_intro}
\input{sec/2_related_work}
\input{sec/3_method}
\input{sec/4_experiments}
\input{sec/5_limitations}
\input{sec/6_conclusion}

{
    \small
    \bibliographystyle{ieeenat_fullname}
    \bibliography{main}
}


\appendix
\input{sec/X_suppl}

\end{document}

%% file: preamble.tex
\usepackage[dvipsnames]{xcolor}

\definecolor{crimson}{rgb}{0.86, 0.08, 0.24}
\definecolor{cadmiumgreen}{rgb}{0.0, 0.42, 0.24}
\definecolor{mediumpersianblue}{rgb}{0.0, 0.4, 0.65}
\definecolor{deepcarrotorange}{rgb}{0.91, 0.41, 0.17}

%% file: sec/0_abstract.tex
\begin{abstract}
The goal of interactive image segmentation is to delineate specific regions within an image via visual or language prompts. Low-latency and high-quality interactive segmentation with diverse prompts remain challenging for existing \textbf{specialist} and \textbf{generalist} models. Specialist models, with their limited prompts and task-specific designs, experience high latency because the image must be recomputed every time the prompt is updated, due to the joint encoding of image and visual prompts. Generalist models, exemplified by the Segment Anything Model (SAM), have recently excelled in prompt diversity and efficiency, lifting image segmentation to the foundation model era. However, for high-quality segmentations, SAM still lags behind state-of-the-art specialist models despite SAM being trained with $\times$100 more segmentation masks. In this work, we delve deep into the architectural differences between the two types of models. We observe that dense representation and fusion of visual prompts are the key design choices contributing to the high segmentation quality of specialist models. In light of this, we reintroduce this dense design into the generalist models, to facilitate the development of generalist models with high segmentation quality. To densely represent diverse visual prompts, we propose to use a dense map to capture five types: clicks, boxes, polygons, scribbles, and masks. Thus, we propose {\SegNext}, a next-generation interactive segmentation approach offering low latency, high quality, and diverse prompt support. Our method outperforms current state-of-the-art methods on HQSeg-44K and DAVIS, both quantitatively and qualitatively.
\end{abstract}

%% file: sec/1_intro.tex
\section{Introduction}
\label{sec:intro}

Interactive image segmentation is a long-standing computer vision task that aims to precisely delineate specific image regions, typically using visual or linguistic prompts. Low-latency interactive segmentation is crucial for prompt decision-making in real-time applications~\cite{wang2005interactive,liu2022isegformer,kirillov2023segment}. With rapid advancements in camera and display technologies, image resolution has been significantly elevated, making 4K and 6K formats increasingly becoming the norm~\cite{shen2022high}. Therefore, the importance of high-quality interactive segmentation for high-resolution images has become more pronounced. On top of these, an effective and practical interactive segmentation approach that supports diverse prompts is key to improving user experience.

\begin{figure}[t]
    \includegraphics[width=0.48\textwidth, height=6.2cm]{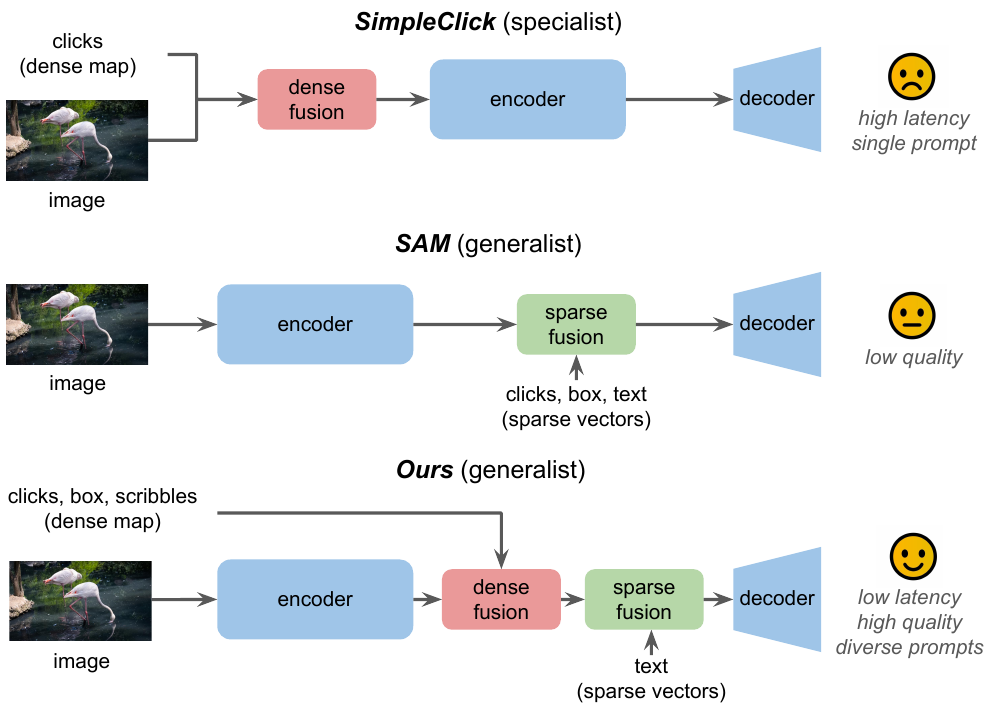}
    \caption{Conceptual comparison between our approach {\SegNext} and prior state-of-the-art methods, SimpleClick~\cite{liu2023simpleclick} and SAM~\cite{kirillov2023segment}, for the interactive segmentation task. Our method combines the best of both worlds for interactive segmentation with low latency, high quality, and diverse prompts.
    }
  \label{fig:teaser}
\end{figure}

However, these requirements remain challenging for existing interactive segmentation approaches, including specialist~\cite{sofiiuk2022reviving,chen2022focalclick,liu2023simpleclick,huang2023interformer} and generalist~\cite{kirillov2023segment,ke2023segment,zhang2023faster,zou2023segment} models. Specialist models, represented by FocalClick~\cite{chen2022focalclick} and SimpleClick~\cite{liu2023simpleclick}, suffer from the high latency issue due to the joint encoding of image and visual prompts. Although InterFormer~\cite{huang2023interformer} circumvents this issue with a decoupled design, it still lacks diversity in the prompts, as only clicks are explored. Generalist models, represented by SAM~\cite{kirillov2023segment} and SEEM~\cite{zou2023segment}, have recently achieved great success in addressing the two issues, lifting image segmentation into the era of foundation models. Nonetheless, in the high-quality segmentation regime, SAM remains a step behind the leading specialist models when more clicks are provided, despite being trained with $\times$100 more segmentation masks. To sum up, designing an optimal architecture for low-latency, high-quality interactive segmentation with diverse prompts remains an unresolved challenge in the field.

To tackle this challenge, we delve deep into the architectural distinctions between specialist and generalist models (Fig.~\ref{fig:teaser} briefly summarizes the major distinctions). We hypothesize that the sparse representation of visual prompts, akin to the treatment of linguistic prompts, could potentially dampen the performance of generalist models for high-quality segmentation. This hypothesis is derived from our answer to a fundamental \textbf{question}: \emph{What are the key distinctions between visual and linguistic prompts in the context of interactive image segmentation?} \textbf{Our answer}: \emph{Visual prompts provide detailed spatial information of image objects, whereas linguistic prompts offer semantic insights without spatial detail. Therefore, linguistic prompts excel in general object recognition, while visual prompts are essential in refining objects for high-quality segmentation.}
However, in generalist models, the sparse representation of these two distinct types of prompts fails to acknowledge their inherent differences; what remains unknown is what specific information is compromised in the representation of visual prompts as sparse vectors. 
A possible explanation suggests that representing visual prompts as sparse vectors, achieved either by interpolating positional embeddings as in SAM~\cite{kirillov2023segment} or by sampling from feature space as in SEEM~\cite{zou2023segment}, amounts to a basic approximation of spatial information. Conversely, a dense map matching the image's spatial dimensions better preserves the detailed spatial attributes of visual prompts. Nevertheless, an in-depth analysis of this issue falls outside the scope of this work.

Driven by our previous analysis, we propose incorporating the dense design, common in specialist models, into generalist models. We propose a three-channel dense map to represent five diverse visual prompts: clicks, boxes, polygons, scribbles, and masks (Sec.~\ref{sec:method_visual_prompts_repr}). This strategy has not yet been extensively explored. We encode the dense map of visual prompts into the image embedding space, followed by element-wise addition and a lightweight fusion module (Sec.~\ref{sec:method_training} and ~\ref{sec:method_impl_details}).

In summary, we propose an approach that can serve as an alternative design of a generalist model for interactive image segmentation with low latency, high quality, and diverse prompts. Extensive evaluations on HQSeg-44K and DAVIS show that our approach outperforms prior state-of-the-art methods both quantitatively and qualitatively. 

%% file: sec/2_related_work.tex
\section{Related Work}
\label{sec:related_work}

\noindent\textbf{Low-latency interactive segmentation.} 
Low-latency segmentation is a field of ongoing research with continuous advancements in model architecture. FCN~\cite{long2015fully} is a seminal work that introduces deep convolutional networks for efficient dense predictions. Originally designed for medical image segmentation, U-Nets~\cite{ronneberger2015u,zhou2018unet++} are highly efficient due to their hierarchical feature representation. While primarily a classification backbone, EfficientNet~\cite{tan2019efficientnet} can be adapted for segmentation. Its scalable architecture makes it a versatile choice for various segmentation tasks. Other advancements include DeepLab~\cite{chen2017deeplab}, HRNet~\cite{wang2020deep}, and ResNet~\cite{he2016deep}. Vision Transformers (ViT) have recently gained popularity in image segmentation. Swin Transformer~\cite{shi2015hierarchical} introduces shifted windows, allowing for efficient local and global feature modeling. SegFormer~\cite{xie2021segformer} uses a hierarchical transformer to reduce computation costs, making it suitable to be deployed on low-end devices. 

Built upon efficient backbones and large-scale datasets, various advanced interactive segmentation approaches have recently been proposed~\cite{sofiiuk2022reviving,chen2022focalclick,liu2022pseudoclick,liu2022isegformer,lin2022knifecut,lin2022focuscut,liu2023simpleclick,lin2020interactive,zhang2020interactive,ding2020phraseclick}. RITM~\cite{sofiiuk2022reviving} adopts HRNet~\cite{wang2020deep} as the backbone and proposes a combination of COCO and LVIS for training, achieving remarkable balance between low latency and state-of-the-art segmentation quality. SimpleClick~\cite{liu2023simpleclick} first uses a large pre-trained plain ViT as the backbone to outperform previous state-of-the-arts such as RITM~\cite{sofiiuk2022reviving} and FocalClick~\cite{chen2022focalclick}. However, SimpleClick suffers from a high-latency issue due to jointly encoding image and visual prompts with the heavy backbone. InterFormer~\cite{huang2023interformer} circumvents this issue by decoupling image encoding and prompt fusion. This decoupled design is also adopted by the Segment Anything Model (SAM)~\cite{kirillov2023segment} for promptable segmentation with low latency. Our method differs significantly from InterFormer and SAM. InterFormer uses multi-scale cross-attention modules to coarsely fuse prompt and image features, while ours uses a much simpler single-scale self-attention module for dense fusion. Besides, InterFormer is a specialist model which does not explore diverse prompts. SAM represents and fuses visual prompts sparsely, while ours uses a dense representation of visual prompts for high-quality segmentation.

\noindent\textbf{High-quality interactive segmentation.}
High-quality segmentation refers to accurately dividing an image into meaningful and distinct segments or regions. It has been explored in various segmentation tasks, including semantic segmentation~\cite{lin2017refinenet,yuan2020segfix}, instance segmentation~\cite{kirillov2020pointrend,zhang2021refinemask}, panoptic segmentation~\cite{de2023intra}, and entity segmentation~\cite{qi2023high}. However, this field remains under-explored for interactive segmentation approaches. This is because high quality and low latency are often at odds, and interactive segmentation approaches must balance the trade-off. Ensuring sufficient quality while maintaining a responsive and real-time experience is crucial for interactive segmentation approaches. Recently, HQ-SAM~\cite{ke2023segment} proposed efficient token learning for accurate mask predictions to improve the segmentation quality of SAM~\cite{kirillov2023segment}. Instead of directly refining a coarse mask as in previous methods, HQ-SAM predicts a new high-quality mask directly by reusing SAM's image encoder and mask decoder. Our method aims not to refine any specific approaches but to reflect on and improve the design choices in existing approaches. Our key intuition is that visual prompts in interactive image segmentation should be treated differently than language prompts, as the two types of prompts have different granularities and semantic meanings. Guided by intuition, we adopt dense representations for visual prompts and sparse ones for linguistic cues, marking the key distinction from HQ-SAM.

\noindent\textbf{Segment anything and beyond.}
The Segment Anything Model (SAM)~\cite{kirillov2023segment} is a recent breakthrough in interactive segmentation. It proposes the Segment Anything Task (SAT) as the ``holy grail'' of interactive image segmentation: a model segments the desired region given a single pixel in the foreground~\cite{li2018interactive}. SAM tackles the ambiguity in this ill-posed task by 1) learning more prior knowledge of object appearance from SA-1B, the largest labeled segmentation dataset, and 2) predicting multiple candidate masks simultaneously. MobileSAM~\cite{zhang2023faster} and HQ-SAM~\cite{ke2023segment} are two SAM variants to achieve more memory-efficient and higher-quality segmentations. With the great success of SAM and its variants, the attention in this field is shifting from \emph{specialist} models to \emph{generalist} models. More recently, SEEM~\cite{zou2023segment} has taken the first step to unify all segmentation tasks in a single model via a universal interface. The key difference between our method and these generalist models lies in the way of representing visual prompts. We adopt a dense representation for visual prompts to keep the spatial information uncompromised. We observe that this dense design leads to high-quality segmentation.

%% file: sec/3_method.tex
\section{SegNext}
\label{sec:method}

\begin{figure*}[t]
    \centering
    \includegraphics[width=16.0cm, height=3.8cm]{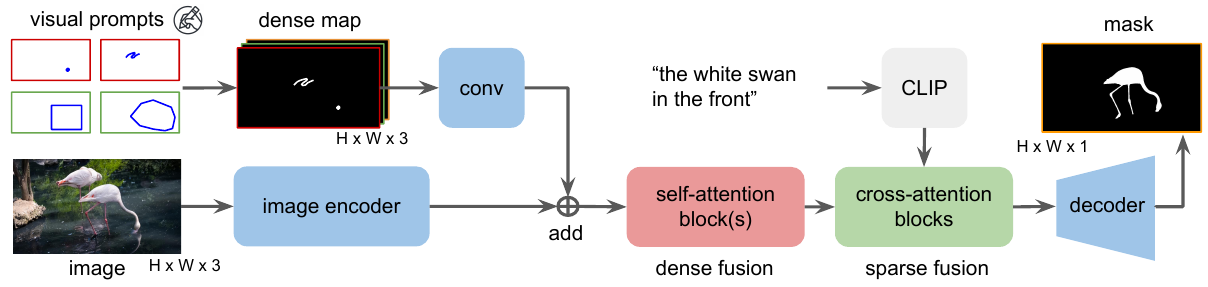}
   \caption{\emph{{\SegNext} overview.} We use a three-channel dense map to represent five diverse visual prompts: clicks, boxes, polygons, scribbles, and masks. The embeddings of image and visual prompts are fused by element-wise addition, followed by an enhanced fusion via one or two self-attention blocks. The language prompt is encoded as a vector by CLIP~\cite{radford2021learning}, followed by querying the image embedding via cross-attention blocks for the mask embeddings. A lightweight decoder processes the mask embeddings for segmentation.}
   \label{fig:method}
\end{figure*}

We propose an interactive image segmentation approach that supports diverse prompts for low-latency and high-quality segmentation. Our method encodes visual and language prompts separately. We use a three-channel dense map to encode five visual prompts: clicks, boxes, scribbles, polygons, and masks. We encode the language prompt as a sparse vector via the CLIP model~\cite{radford2021learning}. To allow for low-latency segmentation, we encode the image \emph{once} to obtain the image embeddings and fuse the prompt features with the image embeddings via lightweight fusion modules. In the following sections, we describe our dense representation strategy for visual prompts (Sec.~\ref{sec:method_visual_prompts_repr}) and how we fuse the visual and language prompts with the image embedding (Sec.~\ref{sec:method_prompts_fusion}). We conclude this section with implementation details for training and inference (Sec.~\ref{sec:method_training} and~\ref{sec:method_impl_details}).

\subsection{Visual Prompts Representation}
\label{sec:method_visual_prompts_repr}

We encode five visual prompts (clicks, boxes, polygons, scribbles, and masks)  via a three-channel dense map with an identical spatial resolution to the input image (\ie, $1024\times1024$). We encode these five visual prompts as follows: 1) \textbf{a click} is encoded as a binary disk with a predefined radius. Positive clicks, which should be put in the foreground, are encoded in the first channel; negative clicks, which should be put in the background, are encoded in the second channel. 2) \textbf{A box} is encoded as four negative clicks, representing the box's four corners. We mandate that the corner points lie outside the foreground area, and we have found that a couple of negative clicks sufficiently convey the object's location. The same logic applies to polygons as well. 3) \textbf{A polygon} is encoded as a set of negative clicks, similar to the box. 4) \textbf{A scribble} is represented by a set of clicks. Similarly to the clicks, a positive scribble is encoded in the first channel, while a negative scribble is encoded in the second. 5) \textbf{A mask} is encoded in the third channel. Unlike the first two channels, the information in this channel can be ambiguous due to false positives or negatives in the initial segmentation mask. Therefore, we encode the mask prompt in a separate channel. The mask prompt allows users to use previous segmentation as the input for the next-round segmentation for better performance.

\subsection{Visual and Language Prompts Fusion}
\label{sec:method_prompts_fusion}

We encode the three-channel dense map via a convolutional layer (\ie, patch embedding layer). The resulting embeddings are element-wisely added to the image embeddings, followed by self-attention blocks for dense fusion. The text prompt can be naturally incorporated after the dense fusion. We use CLIP~\cite{radford2021learning} to encode the text prompt into a vector and query the aforementioned fused embeddings via cross-attention blocks, as introduced in SAM~\cite{kirillov2023segment}. In this work, we primarily concentrate on visual prompts, exploring text prompts only in the appendix as a proof of concept. We remove the text prompt in our main experiments as no language annotations are provided in the main training sets.

\subsection{End-to-end Training}
\label{sec:method_training}

Since clicks can represent all visual prompts, we find that a model trained only with clicks generalizes well to other types of visual prompts. Therefore, we only simulate clicks during training, though it remains unexplored whether training with the other interaction types improves overall performance. Once the model is trained, we can prompt it with ``unseen'' visual prompts, such as boxes and scribbles.

\noindent\textbf{Clicks simulation.} Given the ground truth and current segmentation, we can automatically simulate clicks for training. To boost the performance, we follow RITM~\cite{sofiiuk2022reviving} to use a combination of random and iterative click simulation strategies. The random click simulation strategy generates a set of positive and negative clicks without considering the order between them. In contrast, the iterative simulation strategy generates clicks sequentially: a new click is generated based on the erroneous region of a prediction produced by a model using the set of previous clicks.

\noindent\textbf{Loss function.} We use normalized focal loss~\cite{sofiiuk2022reviving}, allowing faster convergence and better accuracy than binary cross-entropy loss for interactive
segmentation tasks as demonstrated in previous works~\cite{sofiiuk2022reviving,chen2022focalclick}. Similar training pipelines have been proposed by previous works~\cite{sofiiuk2022reviving,chen2022focalclick,liu2023simpleclick}.

\subsection{Implementation Details}
\label{sec:method_impl_details}

\noindent\textbf{Image encoder.} We use ViT-Base (ViT-B/16)~\cite{dosovitskiy2020image} as the image encoder that takes as input an image of size 1024$\times$1024. Following previous works~\cite{liu2023simpleclick,huang2023interformer,kirillov2023segment}, we use readily available pre-trained weights (\ie, MAE~\cite{he2022masked}) to initialize the model for stable and efficient training. The image encoder only encodes each image once. This process can be done offline before prompting the model.

\noindent\textbf{Prompt encoder.} We use two types of prompts: \emph{sparse} (text) and \emph{dense} (clicks, boxes, scribbles, polygons, masks). We represent all dense prompts in a three-channel dense map and free-form text with a CLIP model~\cite{radford2021learning} (\texttt{ViT-L/14@336px}), an off-the-shelf text encoder. Dense prompts (\ie, visual prompts) are encoded via convolutions and added element-wisely with the image embedding.
 
\noindent\textbf{Fusion modules.} After the summation of image embeddings and the embeddings of dense prompts, we use self-attention blocks to enhance the features further. For the sparse prompt (\ie, text), we use the same cross-attention blocks as used in SAM~\cite{kirillov2023segment}.

\noindent\textbf{Decoder.} Our decoder can be any segmentation decoder. In our implementation, we use the same lightweight decoder introduced in SimpleClick~\cite{liu2023simpleclick} for its simplicity and efficiency. It has a simple feature pyramid and two MLP layers. The simple feature pyramid consists of four parallel convolutional layers, to convert the single-scale image feature to a multi-scale one. The MLP layers upsample and combine the multi-scale representation to the output mask, from a multi-scale representation (\ie, \{$\frac{1}{32}$, $\frac{1}{16}$, $\frac{1}{8}$, $\frac{1}{4}$\} of the image resolution) to a single scale representation (\ie, \{$\frac{1}{4}$\} of the image resolution). The output is resized to the image's resolution to obtain the final segmentation mask.

\noindent\textbf{Training and inference settings.} During training and inference, the clicks are encoded as a disk map with a fixed radius of 5 if not otherwise specified. 
We train our models on COCO~\cite{lin2014microsoft}+LVIS~\cite{gupta2019lvis} for 90 epochs with batch size 16 and fine-tune the models on HQSeg-44K~\cite{ke2023segment} for an additional epoch. We observe that finetuning longer does not improve the performance. We use 4 A6000 GPUs for training and 1 A6000 for inference. The initial learning rate is set to $5\times10^{-5}$ and decreases to $5\times10^{-6}$ after epoch 50. We use the following data augmentation strategies: 1) uniform random resizing, 2) random horizontal and vertical flips, 3) random 90-degree rotations, 4) shift scale rotate transforms, and 5) random brightness contrasts. During training, we resize the image's longest side to 1024 and pad the image to 1024$\times$1024. During inference, we directly resize the test image to 1024$\times$1024 without padding.

%% file: sec/4_experiments.tex
\section{Experiments}
\label{sec:experiments}

\begin{figure*}[t!]
        \includegraphics[width=8.8cm, height=6cm, trim=40 5 40 5, clip]{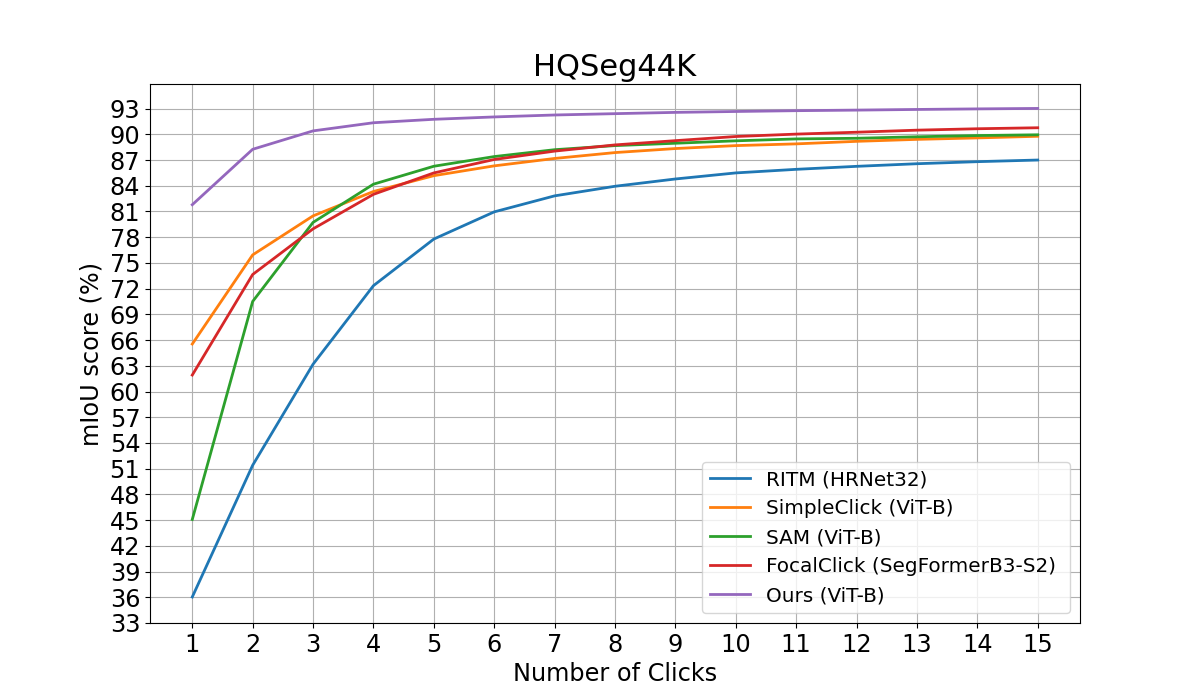}
        \includegraphics[width=8.8cm, height=6cm, trim=40 5 40 5, clip]{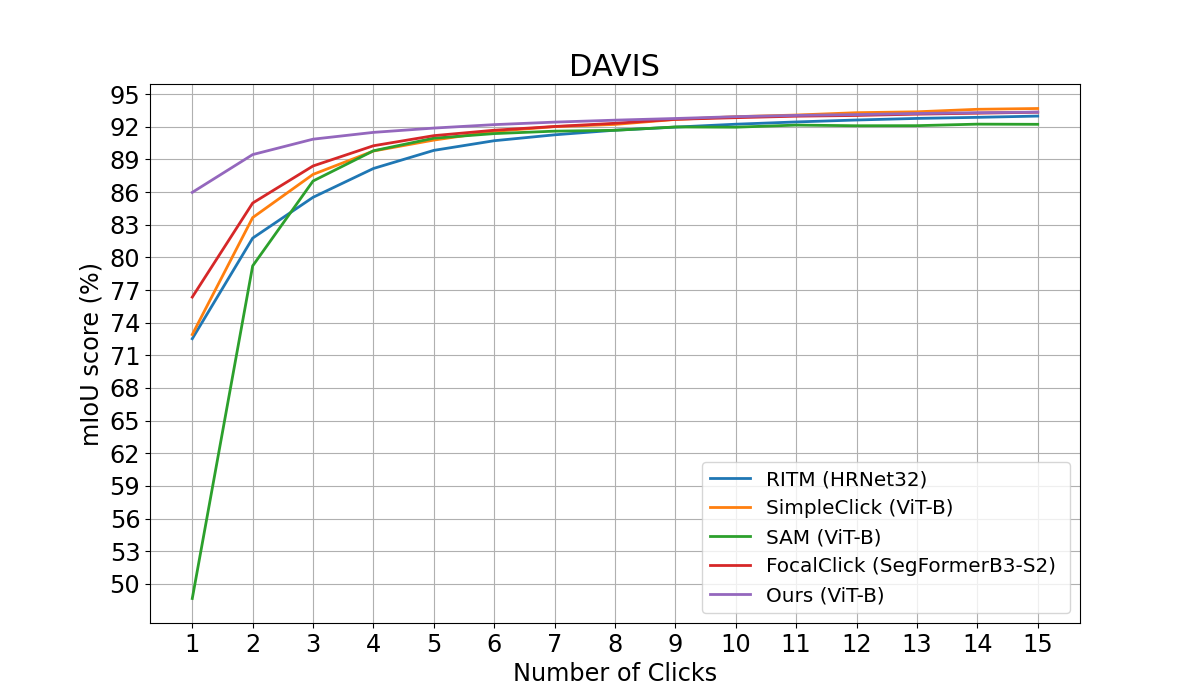}
    \caption{\emph{Click to segmentation evaluation on HQSeg-44K and DAVIS.} With varying numbers of clicks, our method consistently outperforms existing competitive approaches. The metric is mean Intersection over Union (mIoU).}
    \label{fig:convergence_analysis}
\end{figure*}

\begin{table*}
  \centering
  \resizebox{\textwidth}{!}{
  \begin{tabu}[c]{l l c c c c c c c c c c c}
    \toprule
    \multirow{2}{*}{\textbf{Method}} & 
    \multirow{2}{*}{\textbf{Backbone}} &
    \multirow{2}{*}{\textbf{Training}} &
    \multirow{2}{*}{\textbf{SAT}} &
    \multicolumn{4}{c}{\textbf{HQSeg-44K}} & 
    \multicolumn{4}{c}{\textbf{DAVIS}} \\
    \cmidrule(lr){5-8} \cmidrule(lr){9-12}
    & & \textbf{data} & \textbf{Latency (s)}$\downarrow$ & 
    5-mIoU $\uparrow$ & NoC90 $\downarrow$ & NoC95 $\downarrow$ & NoF95 $\downarrow$ &
    5-mIoU $\uparrow$ & NoC90 $\downarrow$ & NoC95 $\downarrow$ & NoF95 $\downarrow$ \\
    \midrule
    \textit{Specialist models} \\
    RITM~\cite{sofiiuk2022reviving}         & HRNet32 $_{400}$   & COCO+LVIS & 22.4 &
    77.72 & 10.01 & 14.58 & 910 & 89.75 & 5.34 & 11.45 & 139 \\
    FocalClick~\cite{chen2022focalclick}    & SegF-B3-S2 $_{256}$   & COCO+LVIS & 36.5 &
    84.63 & 8.12 & 12.63 & 835 & 90.82 & 5.17 & 11.42 & 155 \\
    FocalClick~\cite{chen2022focalclick}    & SegF-B3-S2 $_{384}$   & COCO+LVIS & 51.0 &
    85.45 & 7.03  & 10.74 & 649 & 91.22 & 4.90 & 10.40 & 123 \\    
    SimpleClick~\cite{liu2023simpleclick}   & ViT-B $_{448}$     & COCO+LVIS & 70.5 &
    85.11 & 7.47  & 12.39 & 797 & 90.73 & 5.06 & 10.37 & \textbf{107} \\    
    InterFormer~\cite{huang2023interformer} & ViT-B $_{1024}$     & COCO+LVIS & 24.3 &
    82.62 & 7.17  & 10.77 & 658 & 87.79 & 5.45 & 11.88 & 150 \\
    \midrule
    \textit{Generalist models} \\
    SAM~\cite{kirillov2023segment}          & ViT-B $_{1024}$    & SA-1B    & 7.0 &
    86.16 & 7.46 & 12.42 & 811 & 90.95 & 5.14 & 10.74 & 154 \\
    MobileSAM~\cite{zhang2023faster}        & ViT-T $_{1024}$    & SA-1B    & \textbf{6.6} &
    81.98 & 8.70 & 13.83 & 951 & 89.18 & 5.83 & 12.74 & 196 \\    
    HQ-SAM~\cite{ke2023segment}             & ViT-B $_{1024}$    & SA-1B+HQ & 8.3 &
    89.85 & 6.49 & 10.79 & 671 & 91.77 & 5.26 & \textbf{10.00} & 136 \\
    \midrule
    Ours (SA$\times$1) & ViT-B $_{1024}$    & COCO+LVIS     & 13.3 &
    85.41 & 7.47 & 11.94 & 731 & 90.13 & 5.46 & 13.31 & 177 \\    
    Ours (SA$\times$2)                  & ViT-B $_{1024}$    & COCO+LVIS     & 17.6 &
    85.71 & 7.18 & 11.52 & 700 & 89.85 & 5.34 & 12.80 & 163 \\
    Ours (SA$\times$2)                  & ViT-B $_{1024}$    & COCO+LVIS+HQ     & 17.6 &
    \textbf{91.75} & \textbf{5.32} & \textbf{9.42} & \textbf{583} & \textbf{91.87} & \textbf{4.43} & 10.73 & 123 \\
    \bottomrule
  \end{tabu}}
  \caption{\emph{Quantitative comparison with existing methods on HQSeg-44K and DAVIS.} We compare two types of baselines: specialist and generalist models. Our model achieves comparable performance to the specialist baselines but with significantly lower latency; our model achieves comparable performance to the generalist models in terms of latency and segmentation quality despite being trained with much fewer segmentation data. ``HQ'' denotes the HQSeg-44K dataset; ``SA$\times2$'' denotes the model has two self-attention blocks for dense fusion.
  }
  \label{tab:quantitative_comparison}
\end{table*}

\subsection{Benchmarks}
We use COCO+LVIS, a combination of COCO~\cite{lin2014microsoft} and LVIS~\cite{gupta2019lvis}, for training. We mainly evaluate our method on two public benchmarks: HQSeg-44K~\cite{ke2023segment} and DAVIS~\cite{perazzi2016benchmark}. HQSeg-44K is designed to evaluate the performance of high-quality segmentation. DAVIS contains high-quality annotations and has been widely adopted for evaluating interactive image segmentation performance. We also conduct out-of-domain evaluation on two medical datasets: ssTEM~\cite{gerhard2013segmented} and BraTS~\cite{baid2021rsna}. 

\noindent\textbf{Datasets.} 1) \textbf{COCO+LVIS}. COCO contains 118K training images (1.2M instances); LVIS shares the same images with COCO but has a much higher segmentation quality. 2) \textbf{HQSeg-44K}, a collection of six existing image datasets, including DIS~\cite{qin2022highly} (train set), ThinObject-5K~\cite{liew2021deep} (train set), FSS-1000~\cite{li2020fss}, ECSSD~\cite{shi2015hierarchical}, MSRA10K~\cite{cheng2014global}, and DUT-OMRON~\cite{yang2013saliency}. Each of them contains extremely fine-grained image masks. The average number of masks for each dataset is 7.4K. The training set contains 44320 images; the validation set contains 1537 images. We report evaluation results on the validation set. 3) \textbf{DAVIS}, a high-quality and high-resolution densely annotated video segmentation dataset. Following previous works, we only use 345 frames for evaluation. Our out-of-domain evaluation datasets are: 1) \textbf{ssTEM}~\cite{gerhard2013segmented} contains 20 high-resolution medical images, and 2) \textbf{BraTS}~\cite{baid2021rsna} contains 69 magnetic resonance image (MRI) volumes; we test on the same 369 slices used in SimpleClick~\cite{liu2023simpleclick}.

\noindent\textbf{Evaluation metrics.}
Following previous work, we report both the number of clicks (NoC) and the number of failure cases (NoF) metrics. We also introduce a new metric measuring the latency of Segmentation Anything Task (SAT). Our evaluation metrics include:

\begin{itemize}[leftmargin=*,noitemsep,topsep=0pt]
    \item \textbf{mIoU} measures the average intersection over union (IoU) given a fixed number of consecutive interactions. We use clicks as the default interaction type for this metric. For example, 5-mIoU measures the average IoU given five consecutive clicks. Each click is automatically simulated based on the error of the previous prediction. For the first click, the previous segmentation is empty.
    \item \textbf{NoC} measures the number of clicks required to achieve a predefined IoU. We set two target IoUs: 90\% and 95\%. The corresponding metrics are denoted as NoC\%90 and NoC\%95, respectively. Note that this metric is affected by the maximum number of clicks for evaluation. We set this maximum number to 20. Should a scenario necessitate more clicks than this maximum number, it will be considered a failure case (as further explained below). 
    \item \textbf{NoF} measures the number of failure cases. As defined above, a failure case requires more than 20 clicks to achieve a predefined IoU. For example, NoF90 denotes the number of cases requiring more than 20 clicks to achieve 90\% IoU. 
    \item \textbf{SAT Latency} measures the latency for the Segment Anything Task (SAT). Given a grid of points, we measure how long it takes to prompt the model for SAT. For a fair comparison, we fix the input image size at $1024\times1024$ and prompt the model with a grid of $16\times16$ points. This metric favors generalist models where the image only needs to be encoded once, and multiple prompts amortize the encoding cost. Conventional speed metrics such as seconds per click (SPC) are not set to capture the overall latency of SAT, as they only measure one-prompt latency. The capacity of hardware affects this metric dramatically. In this work, we use an A6000 GPU and an Intel Xeon Gold 6226R CPU to obtain SAT latency results in Tab.~\ref{tab:quantitative_comparison}.
\end{itemize}

\subsection{Baselines}
We compare our method with various interactive segmentation models, including four \emph{specialist} and three \emph{generalist} models, as follows:

\begin{itemize}[leftmargin=*,noitemsep,topsep=0pt]
    \item \textbf{RITM}~\cite{sofiiuk2022reviving} is a light-wight model state-of-the-art model that first proposes using COCO+LVIS for training. The other three specialist models also use this training set. The best-performing RITM model uses HRNet-32~\cite{wang2020deep} as the backbone. We use this model for comparison. 
    \item \textbf{FocalClick}~\cite{chen2022focalclick} uses a coarse-to-fine mechanism to achieve practical interactive image segmentation. FocalClick is the only multi-stage approach that requires global segmentation, local refinement, and progressive merging compared to other baselines. For comparison, we use FocalClick-B3-S2, its best-performing model trained on COCO+LVIS. We set the input size to 256$\times$256 for the segmenter and refiner. A larger input size (\eg 384$\times$384 as reported in Tab.~\ref{tab:quantitative_comparison}) can achieve better performance with more computation. 
    \item \textbf{SimpleClick}~\cite{liu2023simpleclick} uses MAE-pretrained backbones to achieve state-of-the-art performance. We use its ViT-base model trained on COCO+LVIS for comparison. 
    \item \textbf{InterFormer.}~\cite{huang2023interformer} decouples image encoding and prompt fusion for efficient segmentation. We use its ViT-base model trained on COCO+LVIS for comparison. This model was trained with input size $512\times512$, but using $1024\times1024$ for inference achieves better performance.
    \item \textbf{SAM}~\cite{kirillov2023segment} is trained with SA-1B, the largest labeled segmentation dataset so far ($\times100$ more masks than COCO+LVIS); we use its ViT-base model for comparison. \textbf{MobileSAM}~\cite{ke2023segment} is variant of SAM for efficient segmentation on mobile devices; we use its ViT-tiny model for comparison. \textbf{HQ-SAM}~\cite{zou2023segment} is a variant of SAM for high-quality segmentation; we use its ViT-base model.
\end{itemize}

\begin{figure*}[t!]
    \includegraphics[width=5.8cm, height=4.3cm, trim=10 0 20 0, clip]{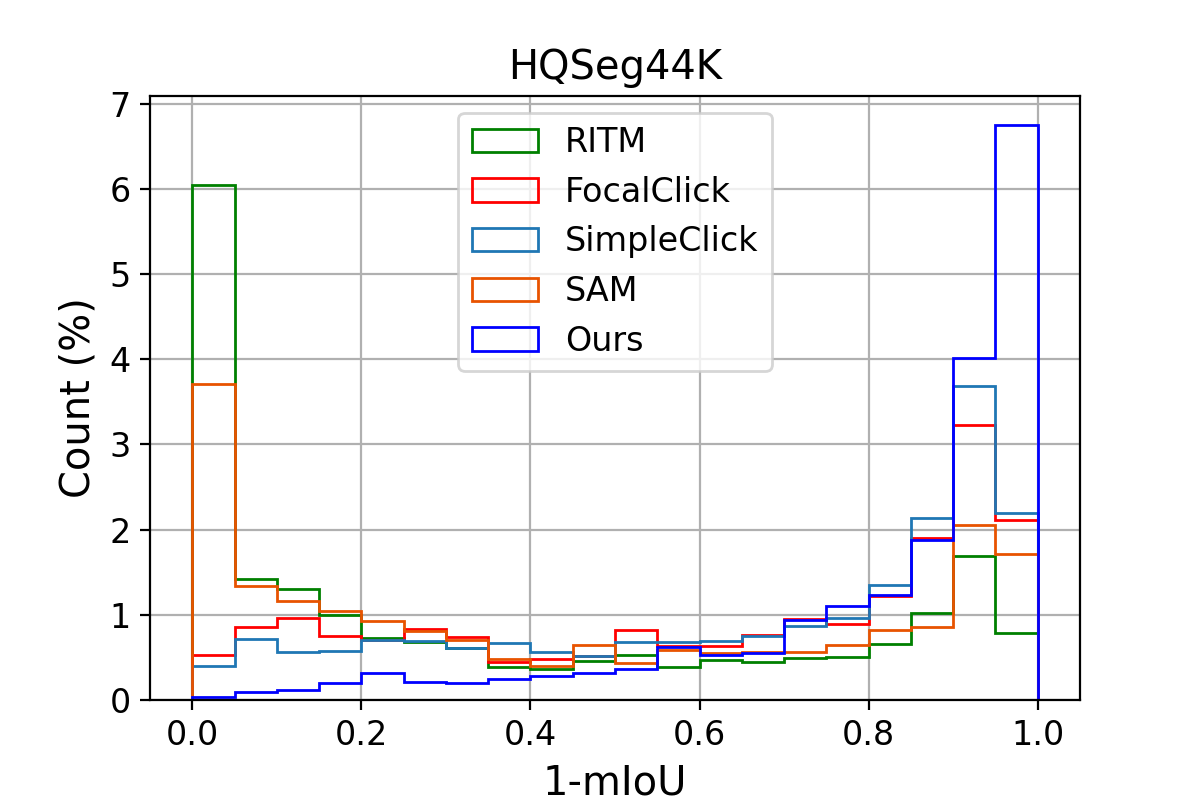}
    \includegraphics[width=5.8cm, height=4.3cm, trim=10 0 20 0, clip]{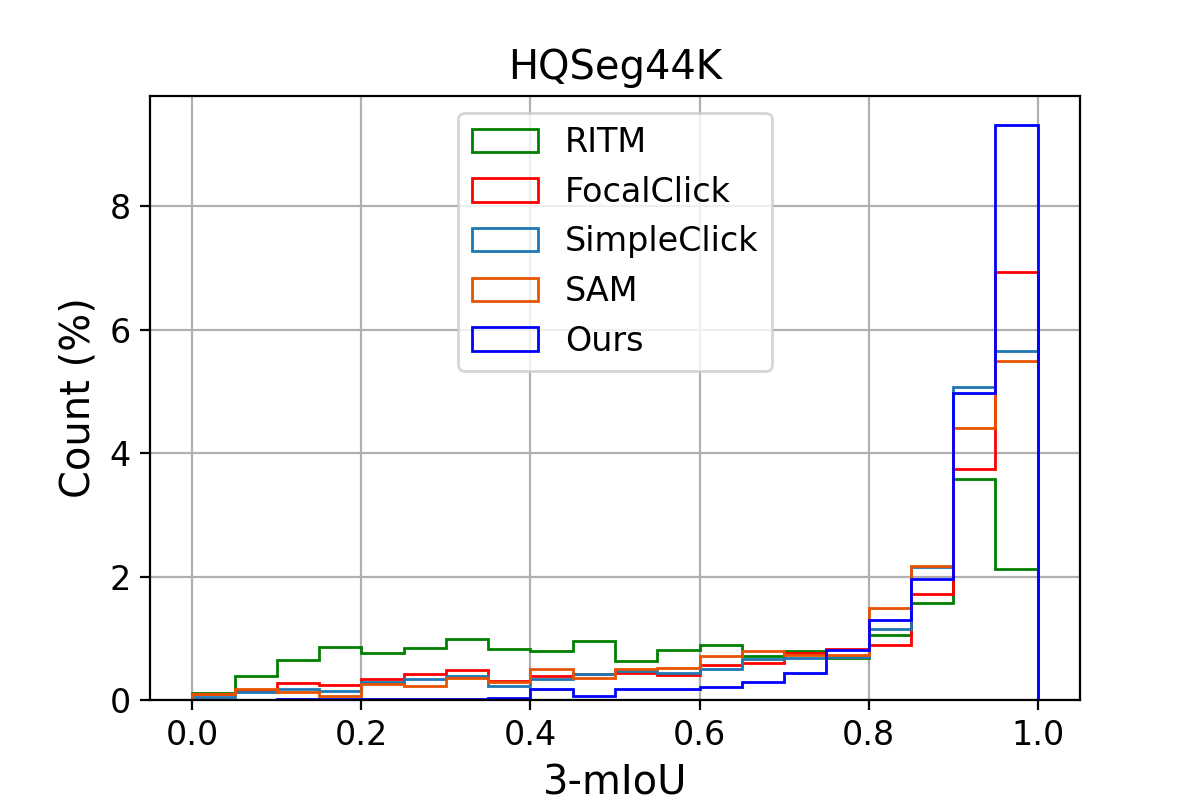}
    \includegraphics[width=5.8cm, height=4.3cm, trim=10 0 20 0, clip]{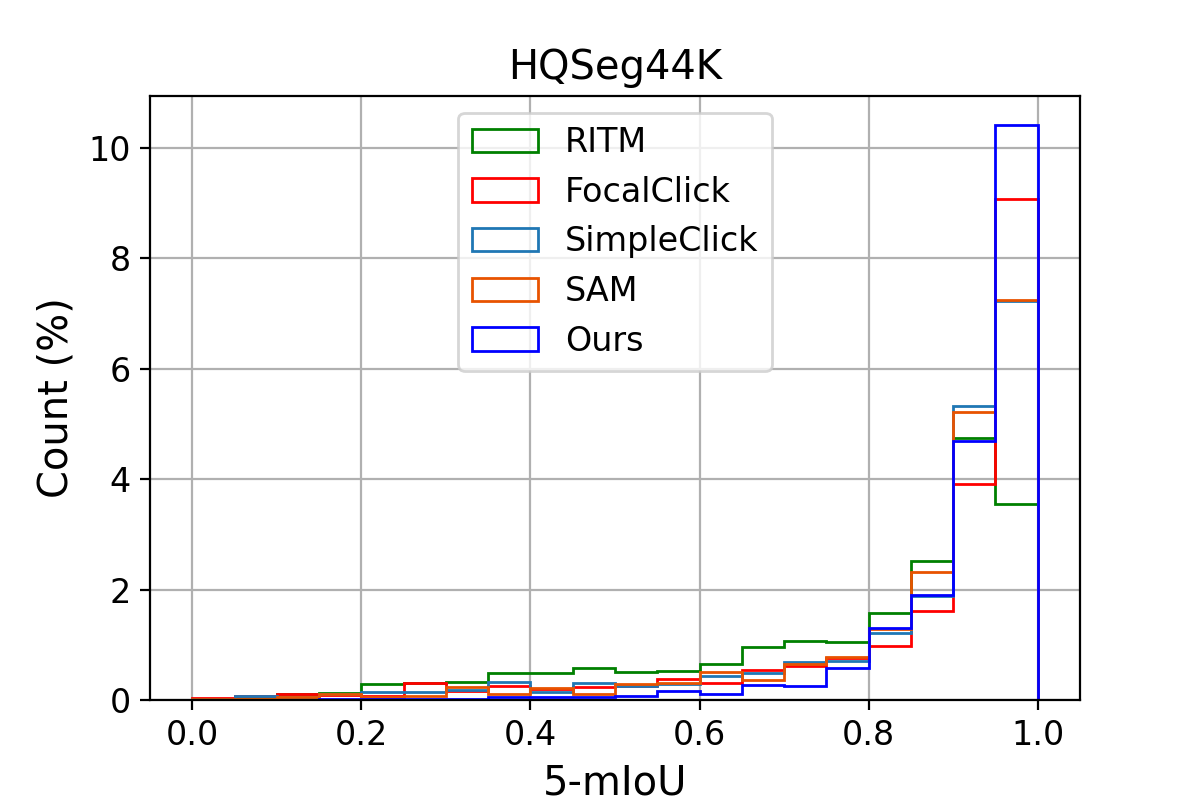}
    \includegraphics[width=5.8cm, height=4.3cm, trim=10 0 20 0, clip]{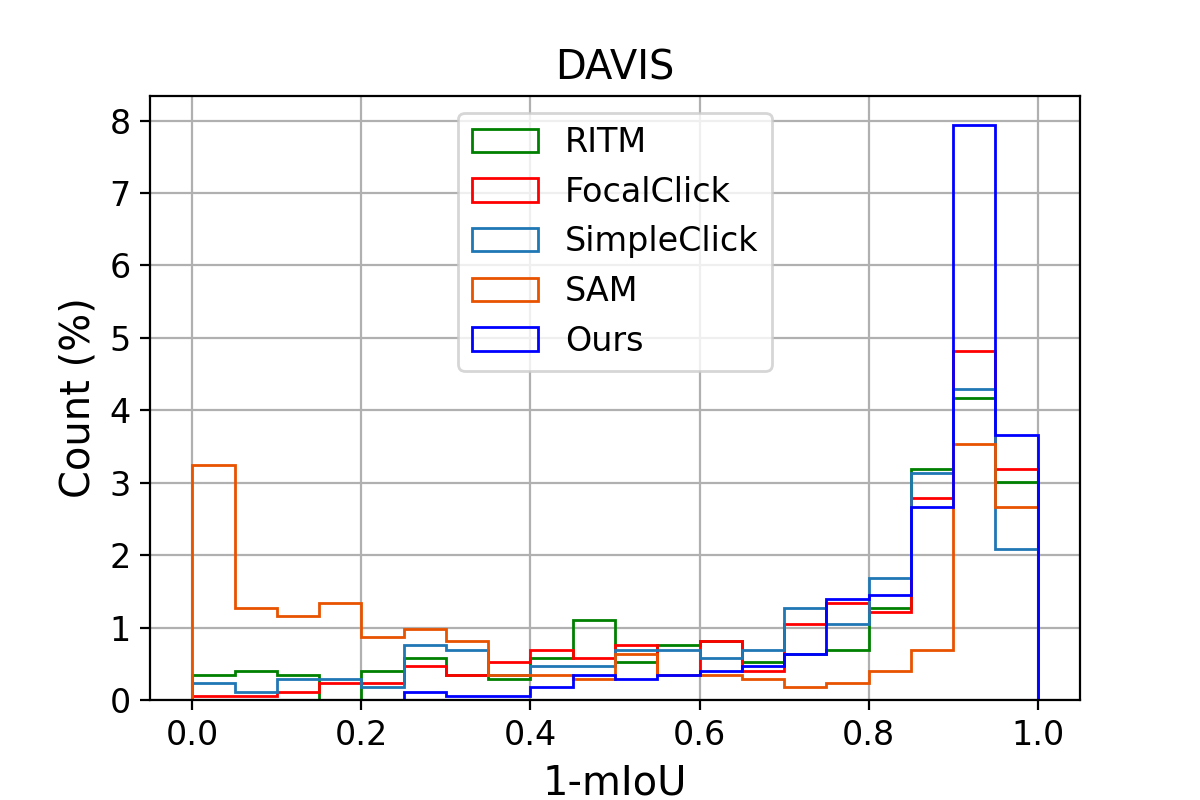}
    \includegraphics[width=5.8cm, height=4.3cm, trim=10 0 20 0, clip]{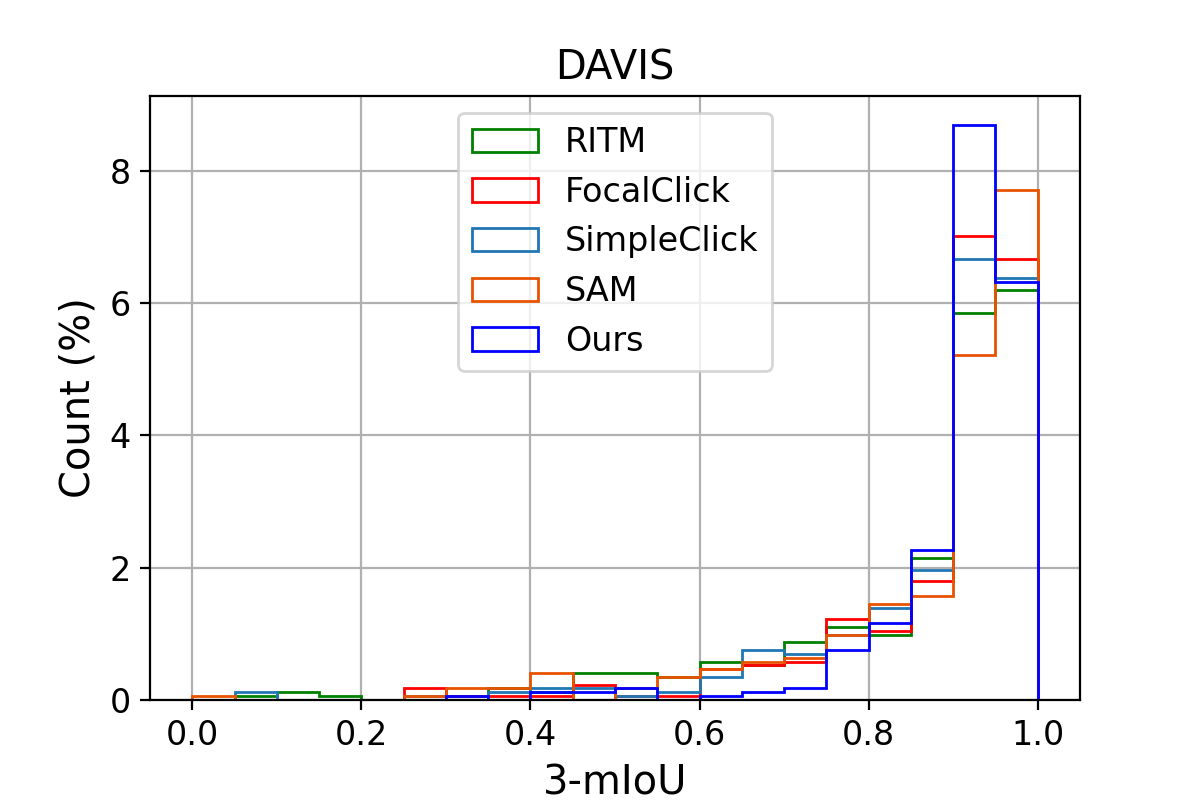}
    \includegraphics[width=5.8cm, height=4.3cm, trim=10 0 20 0, clip]{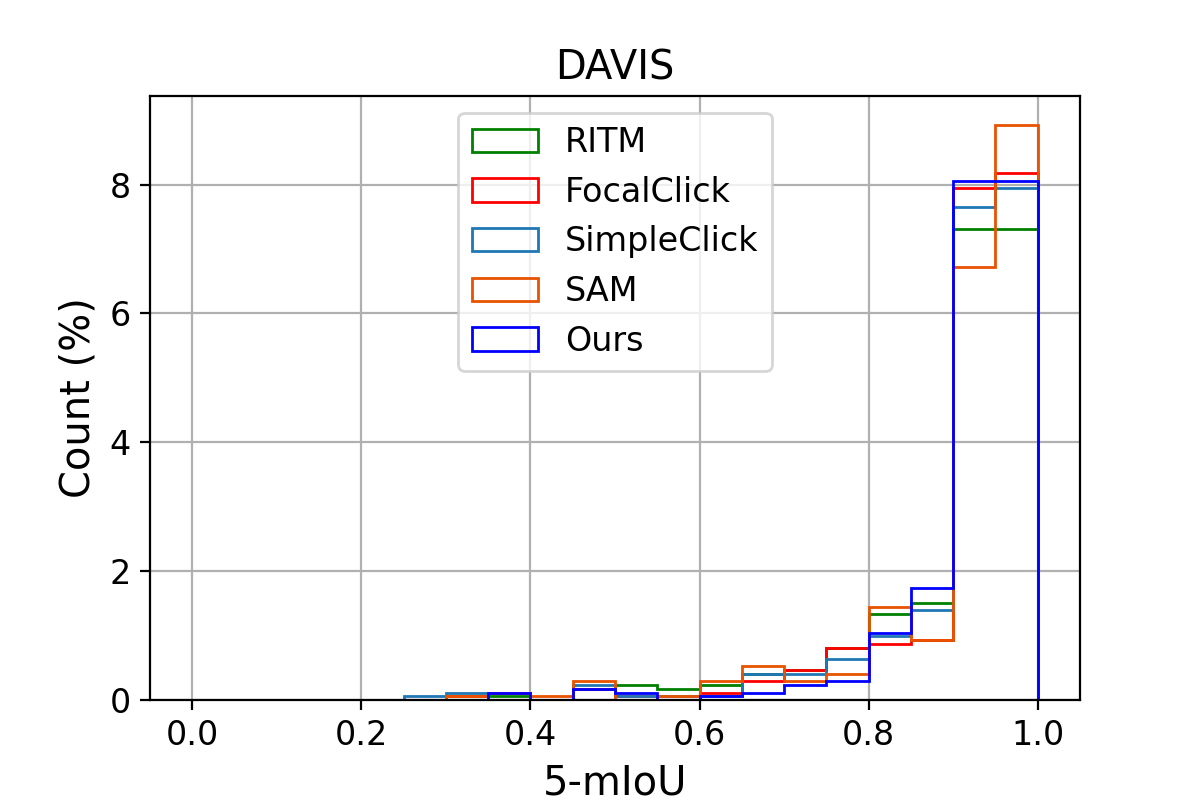}
    \caption{\emph{Histogram analysis} of segmentation IoUs given a predefined $k$ clicks. We report analysis on HQSeg-44K and DAVIS. Compared with the two baselines, our models achieve higher-quality segmentation with fewer failure cases.}
    \label{fig:histogram}
\end{figure*}

\subsection{Comparisons and Analysis}

\noindent\textbf{Quantitative comparisons.} If not otherwise specified, we only use clicks as the default interaction for quantitative comparison. We compare our method with baselines on the HQSeg-44K, and DAVIS benchmarks in Tab.~\ref{tab:quantitative_comparison}. Our method performs best across all segmentation metrics on HQSeg-44K while maintaining low latency. We observe that training on HQSeg-44K significantly boosts our model's segmentation quality. Compared with generalist models, our method performs best while being trained with $\times$100 fewer masks than the mask for training SAM. Compared with specialist models, our method performs competitively while enjoying the lowest latency. Histogram analysis in Fig.~\ref{fig:histogram} shows the distribution of the segmentation quality.

% \noindent\textbf{Qualitative comparisons.} We compare our method qualitatively with our baselines in Fig.~\ref{fig:qualitative_comparison}. We highlight that our method has the highest segmentation quality and the fewest number of failure cases (\ie cases that require more than 20 clicks to achieve 95\% IoU). 

\begin{figure*}[t]
    \centering
    \includegraphics[width=17.5cm, height=4.5cm]{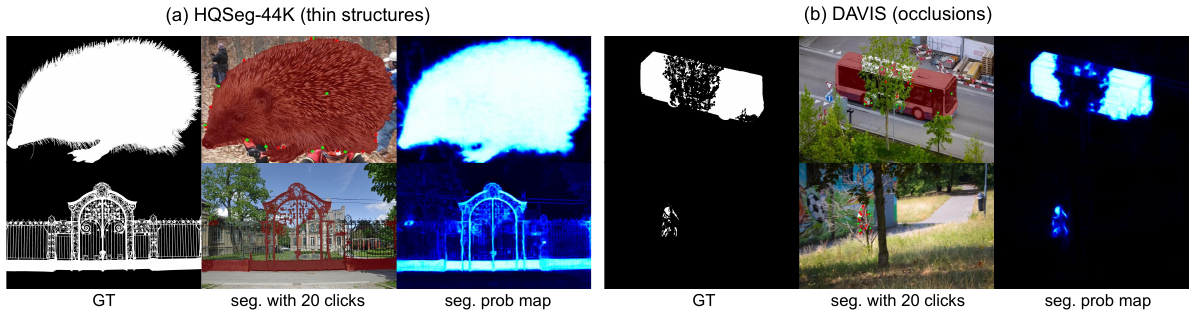}
   \caption{\emph{Failure patterns analysis.} We observed two failure patterns of our method: 1) difficulty with thin structures (as seen on the left) and 2) challenges with cluttered occlusions (as shown on the right). In the case of thin structures, our method tends to overlook fine details of an object, such as missing out on furry textures. In scenarios with cluttered occlusions, our method may struggle to accurately differentiate between the foreground and background.}
   \label{fig:failure_case_analysis}
\end{figure*}

\noindent\textbf{Quantitative analysis} Although our best model achieves superior performance on DAVIS, there are still 123 failure cases among all 345 test cases. We show some failure patterns in Fig.~\ref{fig:failure_case_analysis}.

\subsection{Out-of-Domain Evaluation}
We evaluate the generalizability of our models on two medical image datasets: ssTEM~\cite{gerhard2013segmented} and BraTS~\cite{baid2021rsna}. We compare our method with both specialist and generalist baselines. Our models generalize very well on the two datasets, without the ``zoom-in'' strategy that has been widely adopted in specialist models~\cite{chen2022focalclick,liu2023simpleclick,huang2023interformer,sofiiuk2022reviving,chen2021conditional}. 
``Zoom-in'' is a test-time optimization strategy that focuses on cropping specific local areas, which are determined based on the estimated locations of objects, to achieve finer segmentation. While it is efficient, this approach also incurs extra computational expenses. Tab.~\ref{tab:OOD} reports the evaluation results on the two datasets. Overall, our models generalize well on the out-of-domain datasets, without test-time ``zoom-in''.

\begin{table}
\footnotesize
\centering
\resizebox{0.47\textwidth}{!}{    
\begin{tabu}{l l c c c}
    \toprule
    \multirow{2}{*}{\textbf{Method}} & \multirow{2}{*}{\textbf{Backbone}} & \multirow{2}{*}{\textbf{Zoom-in}} & \textbf{ssTEM} & \textbf{BraTS} \\
    \cmidrule(lr){4-4} \cmidrule(lr){5-5} 
    & & & 10-mIoU $\uparrow$ & 10-mIoU $\uparrow$ \\
    \midrule
    CDN~\cite{chen2021conditional}        &  ResNet-34   & \cmark & 88.46  & 80.24 \\
    RITM~\cite{sofiiuk2022reviving}       &  HRNet32     & \cmark & \textbf{94.11}  & 88.34 \\
    FocalClick~\cite{chen2022focalclick}  &  SegF-B0-S2  & \cmark & 92.62  & 86.02 \\
    FocalClick~\cite{chen2022focalclick}  &  SegF-B3-S2  & \cmark & 93.61  & \textbf{88.62} \\
    SimpleClick~\cite{liu2023simpleclick} &  ViT-B       & \cmark & 93.72  & 86.98 \\
    SAM~\cite{kirillov2023segment}        &  ViT-B       & \xmark & 91.58  & 87.03 \\
    % \rowfont{\color{lightgray}}       
    % SimpleClick~\cite{liu2023simpleclick} &  ViT-L       & \cmark & \textbf{94.34}  & 88.43 \\
    % % \rowfont{\color{lightgray}}       
    % SimpleClick~\cite{liu2023simpleclick} &  ViT-H       & \cmark & 94.08  & \textbf{88.98} \\
    \midrule
    Ours (SA$\times$1) & ViT-B & \xmark & 90.86 & 86.50 \\
    Ours (SA$\times$2) & ViT-B & \xmark & 92.87 & 87.29 \\
    \bottomrule
\end{tabu}}
\caption{\emph{Out-of-domain evaluation} on two medical image datasets: ssTEM~\cite{gerhard2013segmented} and BraTS~\cite{baid2021rsna}. All models are trained on COCO+LVIS. Our models generalize well on the two datasets, without test-time ``zoom-in''. }
\label{tab:OOD}
\end{table}

\begin{figure*}[t]
    \centering
    \includegraphics[width=17.0cm, height=7.0cm]{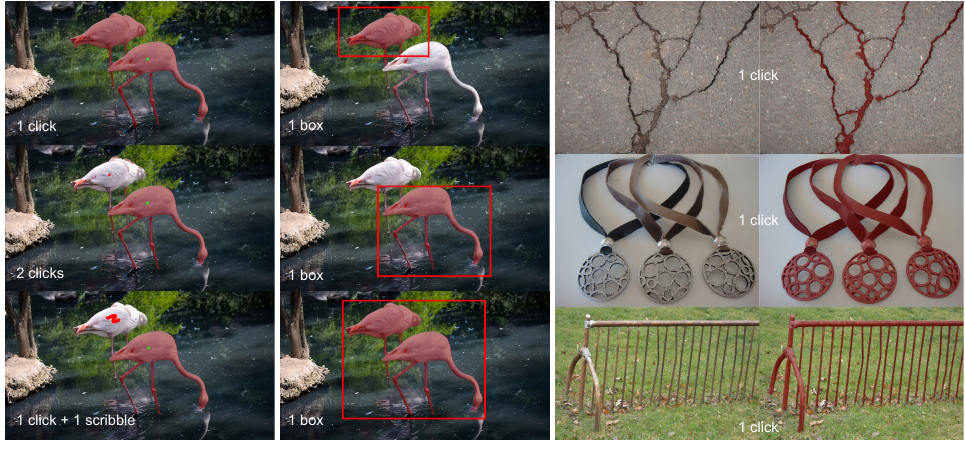}
   \caption{\emph{Qualitative results with diverse prompts.} Left: an example from DAVIS. Right: three examples from HQSeg-44K. The results are achieved by a user providing all the prompts using our best-performing model.}
   \label{fig:diverse_prompts}
\end{figure*}

\subsection{Diverse Prompts Evaluation}

In the previous experiments, we focused solely on using clicks as the interaction mode. However, our method is compatible with a variety of prompts. This section will assess how our model performs with different types of prompts. In Tab.~\ref{tab:diverse_prompts}, we compare our method with two strong baselines, SimpleClick and SAM, on DAVIS. We evaluate four common types of prompts: click, box, scribble, and polygon. The metric ``1-IoU" indicates the mean IoU of all images with a single click. The outcomes highlight our method's robust adaptability, establishing it as a solid foundation for future investigations into various prompts. Fig.~\ref{fig:diverse_prompts} shows qualitative results on DAVIS and HQSeg-44K with diverse prompts. 
The supplementary files also include human evaluation videos to give readers an insight into how our method operates in real applications. These videos provide more details on human evaluations involving natural and medical images.

\begin{table}
\footnotesize
\centering
\resizebox{0.47\textwidth}{!}{
\begin{tabu}{l c c c c}
    \toprule
    \multirow{2}{*}{\textbf{Method}} & \textbf{Click} & \textbf{Box} & \textbf{Scribble} & \textbf{Polygon} \\
    \cmidrule(lr){2-2} \cmidrule(lr){3-3} \cmidrule(lr){4-4} \cmidrule(lr){5-5}
    & 1-mIoU $\uparrow$ & 1-mIoU $\uparrow$ & 1-mIoU $\uparrow$ & 1-mIoU $\uparrow$ \\
    \midrule
    SimpleClick~\cite{liu2023simpleclick} & 72.41 & 69.21  & 76.63 & 68.16 \\
    SAM~\cite{kirillov2023segment}        & 48.66 & 44.87  & 30.42 & 66.49 \\
    \midrule
    Ours (SA$\times$1)                    & 71.17 & 70.37  & 74.70 & 41.33 \\
    Ours (SA$\times$2)                    & 73.61 & 69.08  & 74.26 & 64.04 \\
    \bottomrule
\end{tabu}}
\caption{\emph{Evaluation of diverse prompts on DAVIS.} Although our method is not trained with diverse prompts, it generalizes reasonably well on different ``unseen" visual prompts. We also observe that SAM fails to handle dense visual prompts, such as scribbles. This serves as additional evidence of our hypothesis that dense visual prompts should be represented densely.}
\label{tab:diverse_prompts}
\end{table}

\subsection{Ablations} 

We conduct ablation studies to validate the efficacy of our design choices, as detailed in Tab.~\ref{tab:ablation}. \emph{No dense fusion} is a model variant that removes the self-attention blocks for dense fusion. The prompt embeddings and the image embeddings are fused only by element-wise addition. The segmentation quality of the model drops significantly without the dense fusion module. \emph{No disk} is a model version in which we represent each click as a point on the image instead of a disk. The model's performance drops slightly with this modification. \emph{Weak dense fusion} removes one of the two self-attention blocks for dense fusion. The performance also drops slightly. All the model variants are trained on COCO+LVIS for 90 epochs and tested on HQSeg-44K. 

\begin{table}
    \centering
    \resizebox{0.47\textwidth}{!}{    
    \begin{tabu}[c]{l c c c c}
         \toprule
         \textbf{Method}    & 5-mIoU $\uparrow$ & NoC90 $\downarrow$ & NoC95 $\downarrow$ & NoF95 $\downarrow$ \\
         \midrule
         No dense fusion    & 65.34 & 12.27 & 15.81 & 959 \\
         No disk            & 83.72 & 7.94  & 12.65 & 882  \\
         Weak dense fusion  & 85.41 & 7.47  & 11.94 & 731  \\
         \midrule
         Full               & \textbf{85.71} & \textbf{7.18}  & \textbf{11.52} & \textbf{700}  \\
         \bottomrule
    \end{tabu}}
    \caption{\emph{Ablation study} on HQSeg-44K~\cite{ke2023segment}. \emph{No dense fusion} is a model variant that removes the self-attention blocks for dense fusion. \emph{No disk} is a model version in which we represent each click as a point on the image instead of a disk. \emph{Weak dense fusion} removes one of the two self-attention blocks for dense fusion.} 
    \label{tab:ablation}
\end{table}

%% file: sec/5_limitations.tex
\section{Limitations}
\label{sec:limitations}

Our dense presentation of visual prompts is more costly than a sparse representation used in specialist models. We believe a downsampled dense representation may alleviate this issue. Our text prompt is tentative and not completely stable, although we are confident that it can be enhanced through further dedication. Our method may fail in some challenging scenarios, as shown in Fig.~\ref{fig:failure_case_analysis}. Finally, our work was confined to employing the ViT-Base as the foundational backbone, notwithstanding the availability of more powerful pretrained alternatives such as ViT-Large and ViT-Huge. However, this limitation does not diminish the significance of our findings. Indeed, our results offer valuable insights into the potential for scaling existing models to more substantial backbones or to accommodate larger datasets.

%% file: sec/6_conclusion.tex
\section{Conclusion}
\label{sec:conclusion}

We proposed SegNext for next-generation interactive image segmentation with low latency, high quality, and diverse prompts. We reintroduced the dense representation and fusion of visual prompts, common in specialist models, into the generalist models to facilitate high-quality segmentation. We unified five visual prompts, including clicks, boxes, scribbles, polygons, and masks, on a dense map. We observed that dense representation and fusion of visual prompts are the key design choices contributing to high-quality segmentation. Our method outperformed existing state-of-the-art approaches on challenging benchmarks, both quantitatively and qualitatively.

%% file: sec/X_suppl.tex
% \clearpage
% \setcounter{page}{1}
% \maketitlesupplementary

\section*{Appendix}

\section{Evaluation with the Text Prompt} 
This section supplements the ``Experiments" section in the main paper.
Tab. \ref{tab:text_prompt} quantitatively evaluates the text prompt. The models were trained on RefCOCO~\cite{kazemzadeh2014referitgame} and evaluated on its testA subset across three settings: text-only, click-only, and a combination of text and click (text+click). Following PhraseClick~\cite{ding2020phraseclick}, we used three clicks for the click-only setting and two clicks for the text+click setting. While our text prompt had much room to improve, it yielded promising results combined with visual prompts. 

\begin{table}[h]
  \centering
  \small
  \renewcommand\arraystretch{0.9}
  \begin{tabular}{l c c c c c c}
    \toprule
    Method & Interaction &  Text & Click & mIoU (\%)\\
    \midrule
    PhraseClick~\cite{ding2020phraseclick} & Text-only  & \cmark & \xmark & 50.98 \\
    Ours (SA$\times$2) & Text-only  & \cmark & \xmark & 58.32 \\
    Ours (SA$\times$2) & Click-only & \xmark & \cmark & 82.79 \\
    Ours (SA$\times$2) & Text+Click & \cmark & \cmark & 85.95 \\
    \bottomrule
  \end{tabular}
  \caption{Evaluation with text prompt on the testA of RefCOCO. Our model attained enhanced performance by integrating text and click prompts, surpassing the results achieved with clicks alone.}
  \label{tab:text_prompt}
\end{table}